\documentclass[journal]{IEEEtran}
\usepackage{amsmath,amsfonts}
\usepackage{algorithmic}
\usepackage{algorithm}
\usepackage{multirow}
\usepackage{array}
\usepackage[caption=false,font=normalsize,labelfont=sf,textfont=sf]{subfig}
\usepackage{textcomp}
\usepackage{stfloats}
\usepackage{url}
\usepackage{svg}
\usepackage{hyperref}
\usepackage{verbatim}
\usepackage{graphicx}
\usepackage{cite}
\usepackage{xcolor}
\usepackage{svg}
\usepackage{microtype}

\IEEEoverridecommandlockouts

\usepackage{booktabs}
\begin{document}

\title{DMM: Disparity-guided Multispectral Mamba for Oriented Object Detection in Remote Sensing}

\author{{Minghang Zhou, Tianyu Li*, Chaofan Qiao, Dongyu Xie, Guoqing Wang*,~\IEEEmembership{Member,~IEEE,} Ningjuan Ruan, Lin Mei, Yang Yang,~\IEEEmembership{Senior Member,~IEEE}}  

\thanks{This work was supported in part by the National Natural Science Foundation of China under grant U23B2011, 62102069, U20B2063 and 62220106008, the Key R\&D Program of Zhejiang under grant 2024SSYS0091, the Sichuan Science and Technology Program under Grant 2024NSFTD0034. \textit{(Corresponding author: Guoqing Wang and Tianyu Li.)}}

\thanks{Minghang Zhou, Tianyu Li ,Chaofan Qiao, Dongyu Xie and Guoqing Wang are with the Center for Future Media and School of Computer Science and Engineering, University of Electronic Science and Technology of China, Chengdu 611731, China (e-mail: zminghang@std.uestc.edu.cn; cosm
os.yu@hotmail.com; 202311081534@std.uestc.edu.cn; 202322080339@std.ue
stc.edu.cn; gqwang0420@uestc.edu.cn).} 

\thanks{Ningjuan Ruan is with the Beijing Institute of Space Mechanics and Electricity, Beijing 100094, China.}

\thanks{Lin Mei is with Donghai Laboratory, Zhoushan, Zhejiang 316021, China, and also with the Beijing Institute of Space Mechanics and Electricity, Beijing 100094, China (e-mail: mei\_lin@vip.126.com).}

\thanks{Yang Yang is with the Center for Future Media, and the School of Computer Science and Engineering, University of Electronic Science and Technology of China, Chengdu 611731, China, and also with the Institute of Electronic and Information Engineering, University of Electronic Science and Technology of China, Guangdong 523808, China (e-mial: yang.yang@uestc.edu.cn).}

}

\maketitle

\begin{abstract}
Multispectral oriented object detection faces challenges due to both inter-modal and intra-modal discrepancies. Recent studies often rely on transformer-based models to address these issues and achieve cross-modal fusion detection. However, the quadratic computational complexity of transformers limits their performance. Inspired by the efficiency and lower complexity of Mamba in long sequence tasks, we propose Disparity-guided Multispectral Mamba (DMM), a multispectral oriented object detection framework comprised of a Disparity-guided Cross-modal Fusion Mamba (DCFM) module, a Multi-scale Target-aware Attention (MTA) module, and a Target-Prior Aware (TPA) auxiliary task. The DCFM module leverages disparity information between modalities to adaptively merge features from RGB and IR images, mitigating inter-modal conflicts. The MTA module aims to enhance feature representation by focusing on relevant target regions within the RGB modality, addressing intra-modal variations. The TPA auxiliary task utilizes single-modal labels to guide the optimization of the MTA module, ensuring it focuses on targets and their local context. Extensive experiments on the DroneVehicle and VEDAI datasets demonstrate the effectiveness of our method, which outperforms state-of-the-art methods while maintaining computational efficiency. Code will be available at https://github.com/Another-0/DMM.
\end{abstract}


\section{Introduction} 

Object detection is a fundamental task in the field of remote sensing, with extensive applications in civilian and military areas such as urban planning~\cite{burochin2014detecting, sukel2020urban}, traffic surveillance~\cite{wu2021multi,tao2017object}, disaster relief~\cite{ma2019detection,pi2020convolutional,Merkle_2023_ICCV}, and military reconnaissance~\cite{liu2022hybrid}. However, traditional object detection methods often struggle with the unique challenges posed by remote sensing images~\cite{li2020object}, 
\begin{figure}[!thbp]
    \vspace{+0.1cm}
	\begin{center}
	\includegraphics[width=1\linewidth]{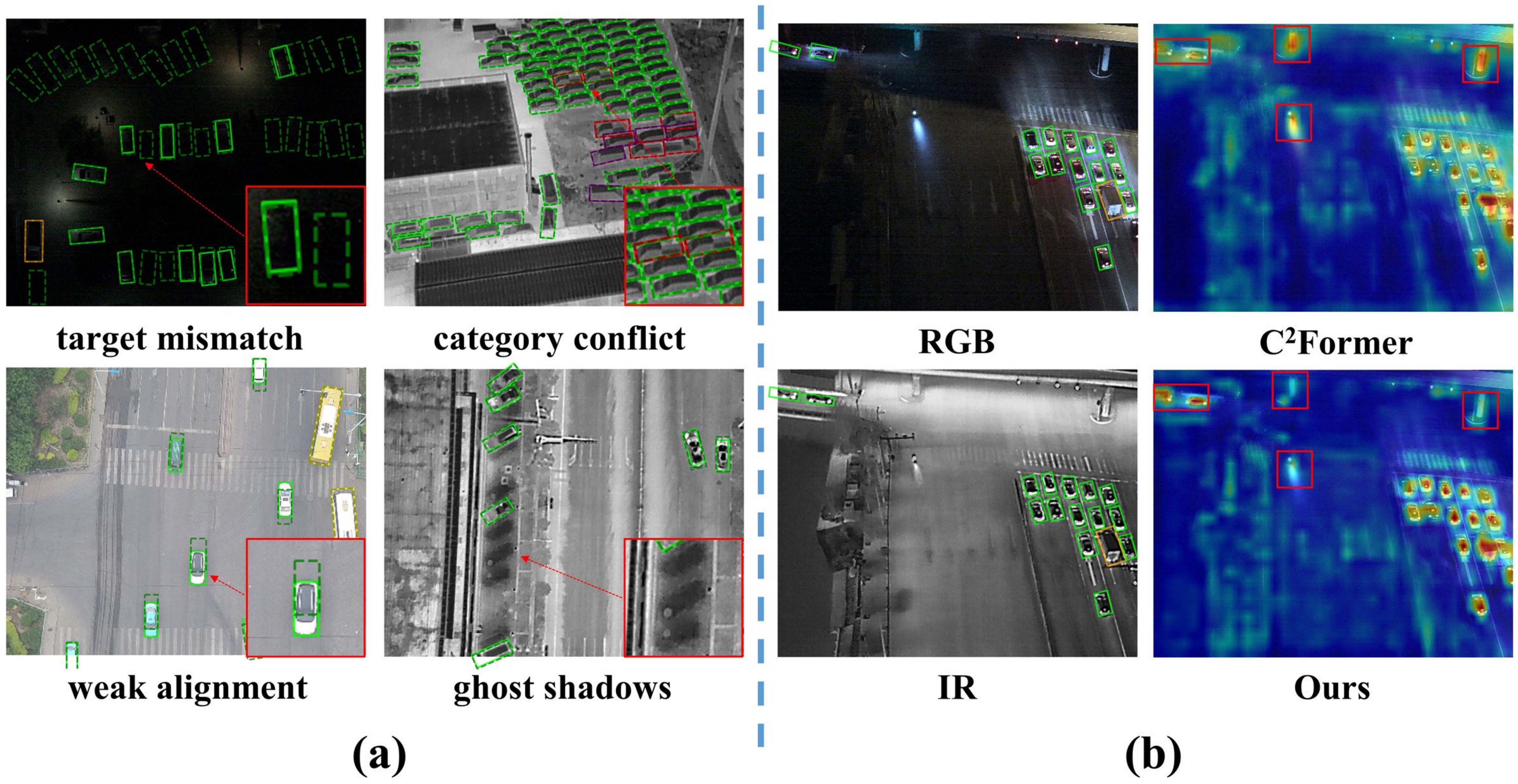}
	\end{center} 
    \vspace{-0.5cm}
    \caption{Modality Disparities between RGB and IR images. Different color bounding boxes represent different categories. Bounding boxes from the RGB modality are represented with solid lines, while dashed lines are from IR modality. (a) shows examples of inter-modal disparities. Target mismatch arises from the varying visibility of targets across different modalities. Category conflict indicates that the differences cause confusion in manual annotation. Due to calibration errors of the capturing devices, paired images are not perfectly aligned. The characteristic of infrared thermal crossover imaging results in ghost shadows in the IR images. (b) shows the intra-modal disparities in RGB images.Uneven illumination can generate a lot of misleading information.C\textsuperscript{2}Former~\cite{yuan2024c} incorrectly focuses on these anomalous regions, whereas our method suppresses background noise and pays more attention to the regions of interest with targets.}
    \vspace{-0.4cm}
    \label{fig:Introduction}
\end{figure}
such as high variability in scale, orientation, and density of objects. To address these challenges, oriented object detection has been developed, which includes the detection of object angles. This approach significantly enhances the accuracy and precision of detecting rotated, irregular, and densely packed objects in complex remote sensing environments~\cite{wang2023oriented}. Consequently, oriented object detection has garnered considerable research attention and application~\cite{ding2019learning,yang2020arbitrary,yang2021r3det,xu2021gliding}. However, these studies are predominantly designed for normal visible(RGB) images, which often suffer from sparse information and significant noise interference due to low illumination and complex weather conditions. 

With advancements in optical sensor technology, infrared (IR) images have been widely adopted as an additional modality to address these challenges, as they can stably reflect the thermal information and contours of objects regardless of illumination and weather~\cite{wei2023unified,sun2022drone}. Although IR images are less impacted by variations in illumination and weather conditions, they offer fewer details about objects, such as color and texture. Therefore, it is intuitive to combine RGB and IR images for object detection, a technique known as multispectral object detection~\cite{Hwang_2015_CVPR,takumi2017multispectral,Xiao_2024_CVPR}.

Based on previous studies~\cite{zhang2021weakly, sun2022drone, yuan2022translation, qingyun2022cross, yuan2024c} and our observations, the challenges for multispectral object detection in remote sensing images can be categorized into two types. One challenge is the inter-modal differences caused by modal characteristics, shooting angles, calibration errors, and post-processing. Fig.~\ref{fig:Introduction}(a) provides some examples of these inter-modal disparities. Another challenge is the substantial intra-modal differences within RGB images, where the image quality of targets can vary dramatically under different lighting conditions, as shown in Fig.~\ref{fig:Introduction}(b). Specifically, artificial light sources at night, such as streetlights, car headlights, and neon signs, can generate misleading information. Additionally, severe noise and shadows caused by extremely low illumination and underexposure interfere with feature extraction, leading to the inclusion of a large amount of irrelevant information in the RGB modality, which hinders optimal fusion in subsequent steps.

Recent studies have applied Convolutional Neural Networks (CNNs) and Transformers to address the fusion challenges. TSFADet~\cite{yuan2022translation} designed an CNN-based alignment framework to solve the weak alignment problem in paired image modalities. However, due to the fixed receptive field size, CNN-based methods struggle to learn global contextual information, which is crucial for detecting numerous small targets in remote sensing images. While Transformer~\cite{vaswani2017attention} have shown effectiveness in capturing long-range dependencies and global information, their expensive computational burden limits its application to high-resolution remote sensing images~\cite{Liu_2021_ICCV}. Consequently, hybrid methods combining CNNs and Transformers have been proposed. For instance, C\textsuperscript{2}Former~\cite{yuan2024c} uses CNNs to extract image features and then employs an Inter-modality Cross-Attention module based on the Transformer architecture to obtain aligned and complementary features, addressing the issue of inaccurate cross-modal fusion. Despite the encouraging results of these methods, they still cannot avoid the quadratic computational complexity of Transformers. Additionally, to reduce computational load, these methods often project features to lower dimensions when computing global attention, which inevitably compromises fusion efficiency. Moreover, existing methods have not adequately explored the differences within the RGB modality and coupled these with inter-modal discrepancies, which are crucial for effective multispectral object detection, particularly under varying illumination conditions.

To address the aforementioned problems, we propose a disparity-guided multispectral oriented object detection framework called DMM. It includes a Disparity-guided Cross-modal Fusion Mamba module (DCFM) to address inter-modal differences, and a Multi-scale Target-aware Attention module (MTA) to handle intra-modal differences within the RGB modality. Benefiting from the selective scanning mechanism and hardware-friendly efficient computation strategies of Mamba~\cite{mamba, mamba2}, DCFM models the global interaction awareness of both single-modality information and modality difference information without sacrificing computational efficiency. MTA employs multi-scale convolutional windows to extract both target and local background information from RGB features, aiming to suppress invalid interference within the modality. Additionally, to guide MTA learn the effective information, a Target-Prior Aware (TPA) auxiliary task is designed, which introduces additional supervisory information through pseudo-labels or manual annotations, using a pre-trained auxiliary detection head to constrain MTA’s optimization process, thus obtaining more beneficial RGB features for subsequent fusion.

Our contributions can be summarized as follows: 
\begin{itemize}
    \item We propose a novel multispectral-oriented object detection framework, DMM, based on the Mamba architecture. Leveraging Mamba’s ability to efficiently capture long-range dependencies, the DCFM module of DMM fuses cross-modal features guided by modality disparity information effectively. To the best of our knowledge, DMM represents the first successful application of Mamba for multispectral-oriented object detection.
    \item We develop the MTA module along with an auxiliary task TPA to bridge the feature gap within RGB modality. TPA introduces additional supervisory information, enabling MTA to focus more on regions where targets are present.
    \item We validate the effectiveness of our proposed method through extensive experiments conducted on two widely used remote sensing datasets, DroneVehicle and VEDAI. Our method significantly outperforms existing state-of-the-art (SOTA) methods, establishing a new benchmark for multispectral-oriented object detection.
\end{itemize}
The rest of the paper is structured as follows: Section II provides a brief overview of related works. Section III introduces the proposed method, and we conduct extensive experiments to validate its effectiveness in Section IV. Finally, we conclude this work in Section V.

\section{Related Work}~\label{sec:relatedwork}

\textbf{Oriented Object Detection.} To extend general object detection methods to rotational scenarios, Yang al.~\cite{yang2020arbitrary}~\cite{yang2021dense} modeled angle prediction as a classification problem, whereas the majority of research focuses on regression-based methods. rotated RPN~\cite{ma2018arbitrary}, RoI Transformer~\cite{ding2018learning} and Oriented R-CNN~\cite{xie2021oriented} propose various strategies to improve the quality of anchor generation for classical anchor-based object detectors. Inspired by ATSS~\cite{zhang2020bridging}, methods such as DAL~\cite{ming2021dynamic}, SASM~\cite{hou2022shape}, GGHL~\cite{huang2022general} and Oriented Reppoints~\cite{li2022oriented} explore the impact of label assignment strategies on oriented object detection performance from different perspectives. To balance accuracy and speed, single-stage based refined detectors, such as R\textsuperscript{3}Det~\cite{yang2021r3det} and S\textsuperscript{2}A-Net~\cite{han2021align} have been proposed. Although these methods have achieved encouraging results in general daytime scenarios, their performance is limited in low-light conditions due to reliance solely on the RGB modality.

\vspace{0.2cm}

\textbf{Multispectral Object Detection.} Introducing the infrared modality in oriented object detection provides robustness against illumination variations~\cite{zhang2019cross, zhang2021guided,li2019illumination,tang2022piafusion}. Numerous studies focus on effectively fusing data from both the RGB and infrared modalities~\cite{huang2022modality, Ye_2021_ICCV, zhou2021ecffnet}. Halfway Fusion~\cite{liu2016multispectral} pioneeringly demonstrated that feature-level fusion yields better results compared to other fusion levels. AR-CNN~\cite{zhang2021weakly} addressed weak alignment discrepancies in multimodal fusion by designing a region feature alignment (RFA) module to capture positional shifts and adaptively align regional features of both modalities. TSFADet~\cite{yuan2022translation} proposed an alignment module to predict the deviation between the two modalities and calibrate the feature maps. The latest method C\textsuperscript{2}Former~\cite{yuan2024c} designed an inter-modal cross-attention (ICA) module to address both modality calibration and fusion inaccuracies, and an Adaptive Feature Sampling (AFS) module to balance the computational cost of global attention. They implemented their modules based on transformer and incorporate numerous strategies to reduce computational load at the expense of fusion accuracy. This prompts us to consider whether it is possible to achieve superior fusion detection results without compromising computational efficiency.

\vspace{0.2cm}

\begin{figure*}[!ht]
    \centering
    \includegraphics[width=1\linewidth]{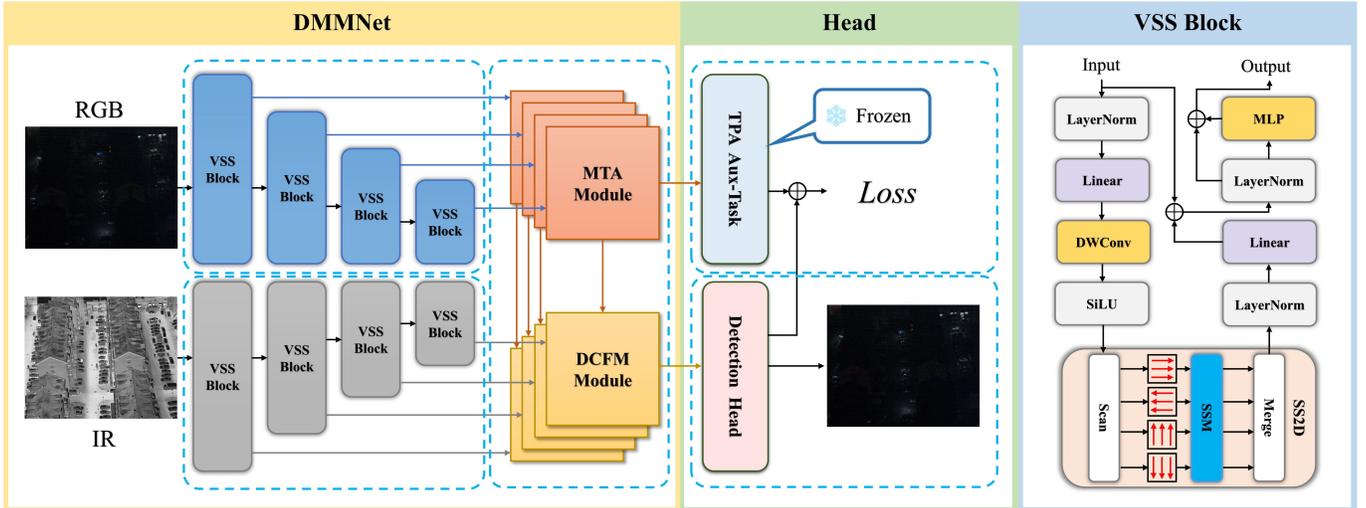}
    \caption{The overview of our proposed DMM method. The input dual-modal features are first projected into a high-dimensional space through convolution, followed by feature extraction using VSS blocks. Each VSS block is cascaded with a downsampling layer to reduce the feature map size. The features of different sizes generated by the VSS blocks in the upper stream and lower stream are fed into the MTA module and the DCFM module, respectively. The output of the MTA is fed into the TPA head to assess the quality of the MTA-enhanced features. The output of the DCFM is directed to the subsequent detection head. Given the critical role of the FPN structure in both one-stage and two-stage algorithms, it is included as part of the detection head architecture. On the far right, we present the structure of the VSS block within the backbone and the SS2D mechanism at the lowest module level, derived from the V9 architecture of VMamba. }
    \label{fig:OverAll}
    \vspace{-0.4cm}
\end{figure*}
\textbf{Mamba.} Mamba is an enhancement of State Space Models (SSMs) ~\cite{gu2021efficiently,mehta2022long, goel2022s,Wang_2023_CVPR} that introduces a selection mechanism and a hardware-aware algorithm, enabling parameterization of SSMs based on the input sequence for efficient processing of long sequences in discrete modalities~\cite{mamba,mamba2}. Recent advances in Mamba have shown significant potential in efficiently handling long sequence modeling~\cite{pioro2024moe,wang2024graph,lieber2024jamba,Tang_2024_CVPR,Nasiri-Sarvi_2024_CVPR}. Mamba offer an alternative to the attention-based models like Transformers by scaling linearly with sequence length and effectively modeling long-range dependencies~\cite{lieber2024jamba, Nasiri-Sarvi_2024_CVPR}. Efforts like Vision Mamba~\cite{zhu2024vision} and VMamba~\cite{liu2024vmamba} extend Mamba's capability to process visual data through bidirectional and multi-directional scanning approaches. Notably, Mamba has also been widely applied in remote sensing tasks~\cite{chen2024changemamba, liu2024rscama,ma2024rs}. RSMamba~\cite{chen2024rsmamba} incorporates a dynamic multi-path activation mechanism to enhance the modeling of non-causal data and demonstrates superior performance in remote sensing image classification. Pan-Mamba~\cite{he2024pan} leverages the Mamba model's efficiency in global information modeling for pan-sharpening, incorporating channel swapping and cross-modal Mamba components to achieve superior fusion results. Samba~\cite{zhu2024samba} employs a unique encoder architecture, based on the Mamba design, to effectively extract semantic information from high-resolution remotely sensed images. Despite the existing research on applying Mamba to general multimodal tasks, its application in multispectral oriented object detection remains unexplored.

\section{Method}\label{sec:method}

This section introduces the proposed Disparity-guided Multispectral Mamba dectection framework, as shown in Fig.~\ref{fig:OverAll}, which consists of three main components: a dual-stream feature extraction backbone network based on a two-dimensional selective state-space model, a multi-scale spatial attention (MTA) module guided by target prior awareness (TPA), and a cross-modal feature fusion module guided by modal disparity (DCFM). We will first introduce the fundamental concepts of the visual selective state-space model in section A. Then, in section B, we will describe the proposed DCFM module. Section C will cover the proposed MTA and TPA modules and their joint optimization process. Finally in section D, we will present the entire oriented object detection process with the integrated detection head and discuss our loss function.

\subsection{Preliminaries}
\textbf{State Space Models (SSMs).} SSMs are a class of models that represent the internal state of a system as a set of linear equations. They are particularly useful for modeling dynamic systems and can be used to capture temporal dependencies in data. Specifically, SSMs can be formulated as linear time-invariant (LTI) systems, which process a one-dimensional input sequence \( x(t) \in \mathbb{R} \) by passing it through intermediate implicit states \( h(t) \in \mathbb{R}^N \) to produce an output \( y(t) \in \mathbb{R} \) where \(N\) is the dimension of the hidden layer. These LTI systems are governed by the following equations and characterized by a set of parameters: the state transition matrix \( \mathbf{A} \in \mathbb{R}^{N \times N} \), the projection parameters \( \mathbf{B} \in \mathbb{R}^{N \times 1}, \mathbf{C} \in \mathbb{R}^{1 \times N} \), and the skip connection \( D \in \mathbb{C}^1 \) :
\begin{equation}\label{eq1}
\begin{split}
  \mathbf{h'}(t) &= \mathbf{A} \mathbf{h}(t) + \mathbf{B} x(t)  \\
  y(t)  &= \mathbf{C} \mathbf{h}(t) + D x(t)
\end{split}
\end{equation}

Due to the continuous-time nature of SSMs, directly applying them in the field of deep learning poses significant challenges, as discretization is required. In the context of computer vision, this means that the continuous state space model must be converted into a discrete form to be compatible with neural network architectures typically used for tasks such as image recognition, object detection, and segmentation. The discretization process involves sampling the continuous-time model at specific intervals, leading to a discrete-time approximation. This transformation is essential for integrating SSMs with standard deep learning frameworks, which operate on discrete data points. Specifically, the discretization of SSMs can be achieved by solving the ODE in Eq.~\ref{eq1}, using methods similar to the zero-order hold principle~\cite{mamba}. For a time scale parameter \( \Delta \), the formulas for converting the continuous parameters
\begin{figure*}[!htp]
    \centering
    \includegraphics[width=1\linewidth]{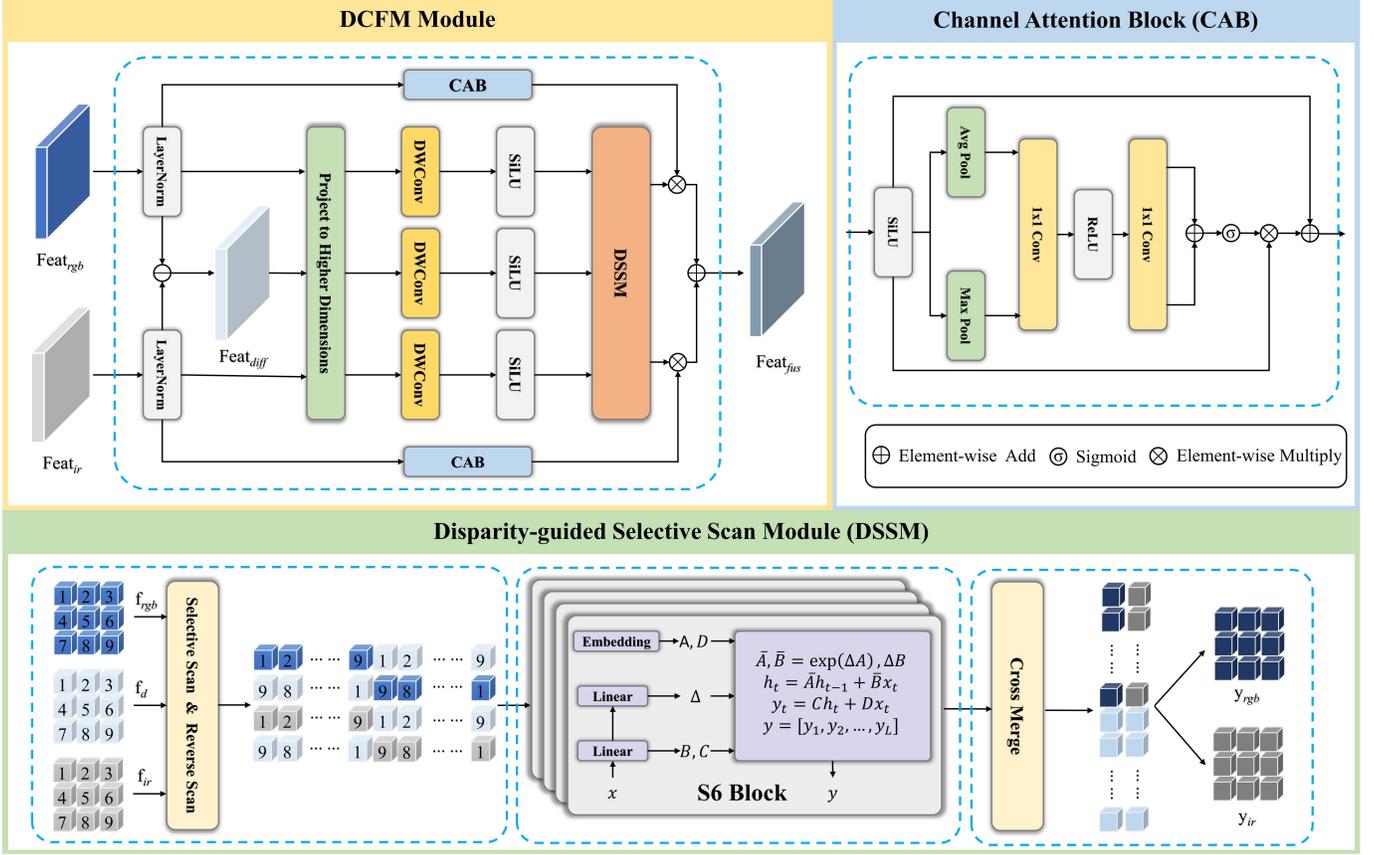}
    \caption{The structures of our proposed DCFM Module. The DCFM Module projects RGB and IR features to higher dimensions, and combines them using DSSM. The Channel Attention Block (CAB) enhances feature representation and the Disparity-guided Selective Scan Module (DSSM) Refines and merges features.}
    \label{fig:dcfm}
    \vspace{-0.4cm}
\end{figure*}
\( \mathbf{A} \) and \( \mathbf{B} \) into their discrete counterparts \( \mathbf{\bar{A}} \) and \( \mathbf{\bar{B}} \) are as follows:
\begin{equation}\label{eq2}
    \begin{split}
        \begin{aligned}
        &\mathbf{\bar{A}}=\exp(\Delta \mathbf{A})\\
        &\mathbf{\bar{B}}=(\Delta \mathbf{A})^{-1}(\exp(\Delta \mathbf{A})-\mathbf{I})\cdot\Delta \mathbf{B}
        \end{aligned}
    \end{split}
\end{equation}

Therefore, Eq.~\ref{eq1} can be rewritten as:
\begin{equation}\label{eq3}
    \begin{split}
        \begin{aligned}
        &h_t=\mathbf{\bar{A}}h_{t-1}+\mathbf{\bar{B}}x_t\\
        &y_t=\mathbf{\bar{C}}h_t+Dx_t \\
        &\bar{C}=C
        \end{aligned}
    \end{split}
\end{equation}

where \( D \) is regarded as a residual connection and is typically omitted in the equation.

\textbf{Selective-Scan Mechanism.} Traditional SSMs, while effective for discrete sequences, face challenges due to their invariant parameters. The Selective State Space Model (S6), also known as Mamba, overcomes this by deriving the matrices \( \mathbf{B} \), \( \mathbf{C} \), and \( \Delta \) from the input data \( x_t \), thus making the model contextually aware~\cite{mamba}. The Selective Scan Mechanism represents a significant advancement in the modeling of complex sequences by introducing input-dependent parameters, addressing the limitations of traditional LTI systems. In computer vision, the direct application of Mamba is challenged by the inherent differences between 2D visual data and 1D sequences. Vision tasks necessitate capturing spatial information, which is not the primary focus in 1D sequence modeling. To bridge this gap, the 2D Selective Scan (SS2D) mechanism was developed by VMamba~\cite{liu2024vmamba}. SS2D arranges image patches in four directions,as showed in Fig.~\ref{fig:OverAll}, generating separate sequences that integrate information from all directions, thereby establishing a global receptive field. This method ensures comprehensive feature integration without increasing 
\begin{figure}[!htbp]
    \centering
    \includegraphics[width=1\linewidth]{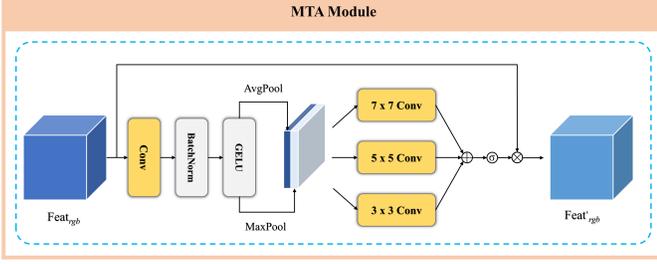}
    \caption{The structure of our proposed MTA Module. The outputs are fed to both the DCFM module for high-precision fusion and the TPA module for loss calculation.}
    \label{fig:mta}
    \vspace{-0.4cm}
\end{figure}
computational complexity. The SS2D mechanism processes each directionally scanned sequence with the S6 model, combining the resultant features to reconstruct the 2D feature map. This approach allows each element in the 1D array to interact with previously scanned samples through a condensed hidden state, effectively capturing contextual information across the entire image.

\subsection{Disparity-guided Cross-modal Fusion Mamba}
The feature fusion module is the core of multimodal models. Existing effective methods primarily apply the dynamic modeling capabilities of transformers to address inter-modal differences. However, their expensive computational cost needs careful consideration. Unlike previous approaches, our DCFM module is based on Mamba, achieving global attention while avoiding quadratic computational overhead. The structure of our proposed DCFM module is shown in Fig.~\ref{fig:dcfm}. Considering a set of inputs \( Feat_{rgb}, Feat_{ir} \in \mathbb{R}^{N \times C \times H \times W} \) representing RGB and IR features respectively, we first normalize the features using a LayerNorm block to accelerate model convergence. Then we calculate the inter-modal disparities and project all of them to unified hidden space: 
\begin{equation}\label{eq4}
    \begin{split}
        \begin{aligned}
            &F_{rgb}=LN(Feat_{rgb}) \\
            &F_{ir}=LN(Feat_{ir}) \\
            &F_{d}= F_{rgb} - F_{ir} \\
            &F^{\prime}_i=Project(F_i), F_i=F_{rgb},F_{ir},F_{d}
        \end{aligned}
    \end{split}
\end{equation}

where \( LN(\cdot) \) represents LayerNorm opreration and \( Project(\cdot) \) represents the projection operation using a linear transformation. Unlike the common dimensionality reduction operations in transformers, the channel dimension of \( X^{\prime} \in \mathbb{R}^{N \times C^{\prime} \times H \times W} \) is typically twice that of the initial dimension(\(C^{\prime} = 2C\)), providing better modality information representation. Subsequently, depthwise separable convolutions \( DWConv(\cdot) \) are applied to the three features to facilitate inter-channel communication:
\begin{equation}\label{eq5}
    \begin{split}
        \begin{aligned}
            f_{i} = SiLU(DWConv(F^{\prime}_{i})), F^{\prime}_{i}=F^{\prime}_{rgb},F^{\prime}_{ir},F^{\prime}_{d}
        \end{aligned}
    \end{split}
\end{equation}

where \( SiLU(\cdot) \) represents SiLU activation function. These features, fully integrated with information from all channels, undergo further processing through the Disparity-guided Selective Scan Module(DSSM) module. This module extracts complementary information from another modality while suppressing redundant data. The out features are then element-wise multiplied and summed with the original features processed by the Channel Attention Block(CAB), ultimately yielding high-quality fused features.

\textbf{ Channel Attention Block.} Although SSM excels at addressing long-range dependencies, it has limitations in modeling inter-channel relationships. To overcome this, we propose CAB to compute channel attention on the normalized original features, adaptively learning the intra-modal channel relationships to enhance the feature representation of a single modality. Specifically, for a given input feature \( F_{i} \), we compute the channel attention as follows:
\begin{equation}\label{eq6}
    \begin{split}
        \begin{aligned}
            &F_{avg} = AvgPool(SiLU(F_i)) \\
            &F_{max} = MaxPool(SiLU(F_i)) \\
            &w = \mathcal{S} (CR(F_{avg})+CR(F_{max})) \\
            &F_{out} = w \cdot F_i +  F_i
        \end{aligned}
    \end{split}
\end{equation}

here, \( CR(\cdot) \) denotes the operation of a 1x1 convolution followed by a ReLU activation function, and \( \mathcal{S}(\cdot) \) represents the Sigmoid function. Both \( AvgPool(\cdot)\) and \( MaxPool(\cdot)\)  are global operations, producing \( F_{avg} \) and \( F_{max} \) in \( \mathbb{R}^{N \times C \times 1 \times 1} \).

\textbf{Disparity-guided Selective Scan Module.}  The unique selective scan mechanism in the Mamba architecture allows it to adjust parameters selectively based on the input data. Leveraging this context-aware capability, we designed the DSSM module. Specifically, the DSSM processes three inputs, \( f_{rgb}, f_{ir}, f_{d} \in \mathbb{R}^{N \times D \times H \times W} \). First, we flatten the features and then concatenate \( f_{d} \) with \( f_{rgb} \) and \( f_{ir} \) to obtain two input sequences \( f_{rgb\text{-}d},f_{ir\text{-}d} \in \mathbb{R}^{N \times 2HW \times D} \). Considering the two-dimensional nature of image features, we reverse scan the sequences to obtain two new input sequences, \( \overline{f}_{rgb\text{-}d} \) and \( \overline{f}_{ir\text{-}d} \). These four sequences are then processed using Eq.~\ref{eq2} and Eq.~\ref{eq3} to produce output sequences \( f'_{rgb\text{-}d}, f'_{ir\text{-}d}, \overline{f}'_{rgb\text{-}d}, \overline{f}'_{ir\text{-}d} \in \mathbb{R}^{N \times 2HW \times D} \). Finally, these four output sequences are applied to the following formula:
\begin{equation}\label{eq7}
    \begin{split}
        \begin{aligned}
            &y_{rgb\text{-}d} = f'_{rgb\text{-}d} + Reverse(\overline{f}'_{rgb\text{-}d}) \\
            &y_{ir\text{-}d} = f'_{ir\text{-}d} + Reverse(\overline{f}'_{ir\text{-}d}) \\
        \end{aligned}
    \end{split}
\end{equation}

where \( Reverse( \cdot ) \) function reverses the input along the second dimension. We then perform the inverse operation of concatenation by splitting \( y_{rgb\text{-}d} \) and \( y_{ir\text{-}d} \) along the same dimension. Retaining the modality-specific features from the first half and discarding the differential features from the second half, we reshape them into the final output features \( y_{rgb\text{-}d} \) and \( y_{ir\text{-}d} \in \mathbb{R}^{N \times D \times H \times W} \).

\subsection{Multi-Scale Target-Aware Attention}
Although the DCFM module effectively integrates multi-modal information, enhancing the representation and understanding of complex visual data, intra-modal variations such as lighting can introduce interference and redundancy in RGB features, as shown in Fig.~\ref{fig:Introduction}(b). This interference hinders the fusion module's ability to effectively distinguish between relevant targets and background noise, thus compromising fusion performance. To address this issue, we propose the Multi-Scale Target-Aware Attention(MTA) module, depicted in Fig.~\ref{fig:mta}. The MTA module computes multi-scale spatial self-attention to adaptively focus on target regions within the RGB modality, thereby providing higher quality RGB features for subsequent fusion modules.

Specifically, for a single-layer feature \( x_{rgb} \in \mathbb{R}^{N \times C \times H \times W} \) output by the VSS block, we first apply a convolution layer for initial processing. We then use channel-wise \( AvgPool(\cdot) \) and \( MaxPool(\cdot) \) operations to aggregate channel information. Finally, we compute global spatial attention weights using convolution operations with different kernel sizes. These weights are used to reweight the original input features, which are then combined with a residual connection to produce the enhanced features.

\textbf{Target-Prior Aware Auxiliary Task.} From our observation, the MTA module alone is insufficient to ensure that the reweighted features focus more on the targets and suppress background noise. To address this, we introduce the Target-Prior Aware (TPA) auxiliary task by adding a TPA head after the MTA module. The structure of TPA is similar to the FPN and RPN layers in traditional single-stream object detectors, but with the parameters frozen during training. We first obtain the TPA weights by pre-training an RGB single-modal two-stage object detector, using either pseudo labels generated by other SOTA single-modal detectors or manually annotated labels. Then We pre-train a detector that includes only the DCFM module to enable the model to learn robust representations from dual-modal images. Finally, we incorporate the MTA and TPA modules into the model for joint training. During the training of our model, the features output by the MTA module are simultaneously fed to the TPA detection head, as shown in Fig.~\ref{fig:OverAll}. The RPN layer's role in extracting regions of interest aligns with our optimization goal, so we use the RPN loss as the optimization target for the auxiliary task. With the TPA auxiliary task constraint, the MTA module is optimized to focus on target regions identified by the RPN, resulting in higher quality RGB features for subsequent fusion modules. The effectiveness of this strategy is clearly illustrated in Section IV-C through ablation experiments.

\subsection{Dual-Stream Detector}
As described earlier, our DMM is a dual-stream feature extraction and fusion network that operates independently of the subsequent detection head. This means that our model can be integrated into any existing SOTA object detection methods simply by replacing the feature extraction architecture. In this paper, we adopt the one-stage method S\textsuperscript{2}A-Net~\cite{han2021align} as our basic detector, adhering to the same configurations as previous SOTA models. To facilitate experimental comparisons and verify generalization, we also integrate our method into the classic two-stage methods Faster R-CNN~\cite{NIPS2015_14bfa6bb} and Oriented R-CNN~\cite{Xie_2021_ICCV} by replacing the feature extraction module. All the detectors use Oriented Bounding Box (OBB) detection heads. We retain their unique detection head designs without any modifications to ensure fair comparisons with other methods.

\textbf{Loss Function.} We adhere to the design of other detection networks, retaining their inherent loss functions. The distinction lies in augmenting these with the loss functions from our TPA auxiliary task. Therefore, the final loss function is defined as follows:
\begin{equation}\label{eq8}
    \begin{split}
        \begin{aligned}
            \textit{Loss}_{All} =&  \textit{Loss}_{\text{det\_cls}} + \textit{Loss}_{\text{det\_reg}} \\
            &+ \textit{Loss}_{\text{aux\_cls}} + \textit{Loss}_{\text{aux\_reg}}
        \end{aligned}
    \end{split}
\end{equation}

where \( \textit{Loss}_{\text{det\_cls}} \) and \( \textit{Loss}_{\text{det\_reg}} \) represent the classification and regression losses of the detection head, respectively, while \( \textit{Loss}_{\text{aux\_cls}} \) and \( \textit{Loss}_{\text{aux\_reg}} \) denote the classification and regression losses of the TPA auxiliary task. All classification loss functions use cross-entropy loss~\cite{6773024}, with binary classification for the TPA auxiliary task; all regression losses use smooth \( L_1 \) loss~\cite{Girshick_2015_ICCV} function. For two-stage detection algorithms, the first two terms include both the RPN network losses and the ROI network losses.

\section{Experiments}\label{sec:experiments}
\subsection{Implementation Details}  
We train and validate our model on the following two datasets:

{\bf DroneVehicle Dataset:} The DroneVehicle dataset~\cite{sun2022drone} is a large-scale remote sensing dataset collected from drones, comprising 28,439 pairs of RGB and infrared images. Each image is annotated with oriented bounding boxes for five categories (car, bus, truck, van and freight car), totaling 953,087 instances. The scenes range from daytime to nighttime, including roads, urban areas, parking lots and etc. The dataset is officially divided into training, validation, and test sets with 17,990, 1,469, and 8,980 image pairs, respectively. During training, we remove the 100-pixel-wide white borders around all images and conduct comparative experiments on the test sets.

{\bf VEDAI Dataset:} The VEDAI dataset~\cite{razakarivony2016vehicle} is designed for vehicle detection in high-resolution aerial images. It includes diverse urban and rural scenes with oriented annotations for various vehicle types such as cars, trucks, and vans, along with a few other objects like airplanes and ships, totaling 9 categories. The dataset consists of 1,246 pairs of RGB and infrared images with resolutions of 1024x1024 and 512x512 pixels. In our experiments, we use the 512x512 version and follow the ten-fold cross-validation protocol as described in \cite{razakarivony2016vehicle} to train and test our model.

\begin{figure*}[!htp]
    \centering
    \includegraphics[width=1\linewidth]{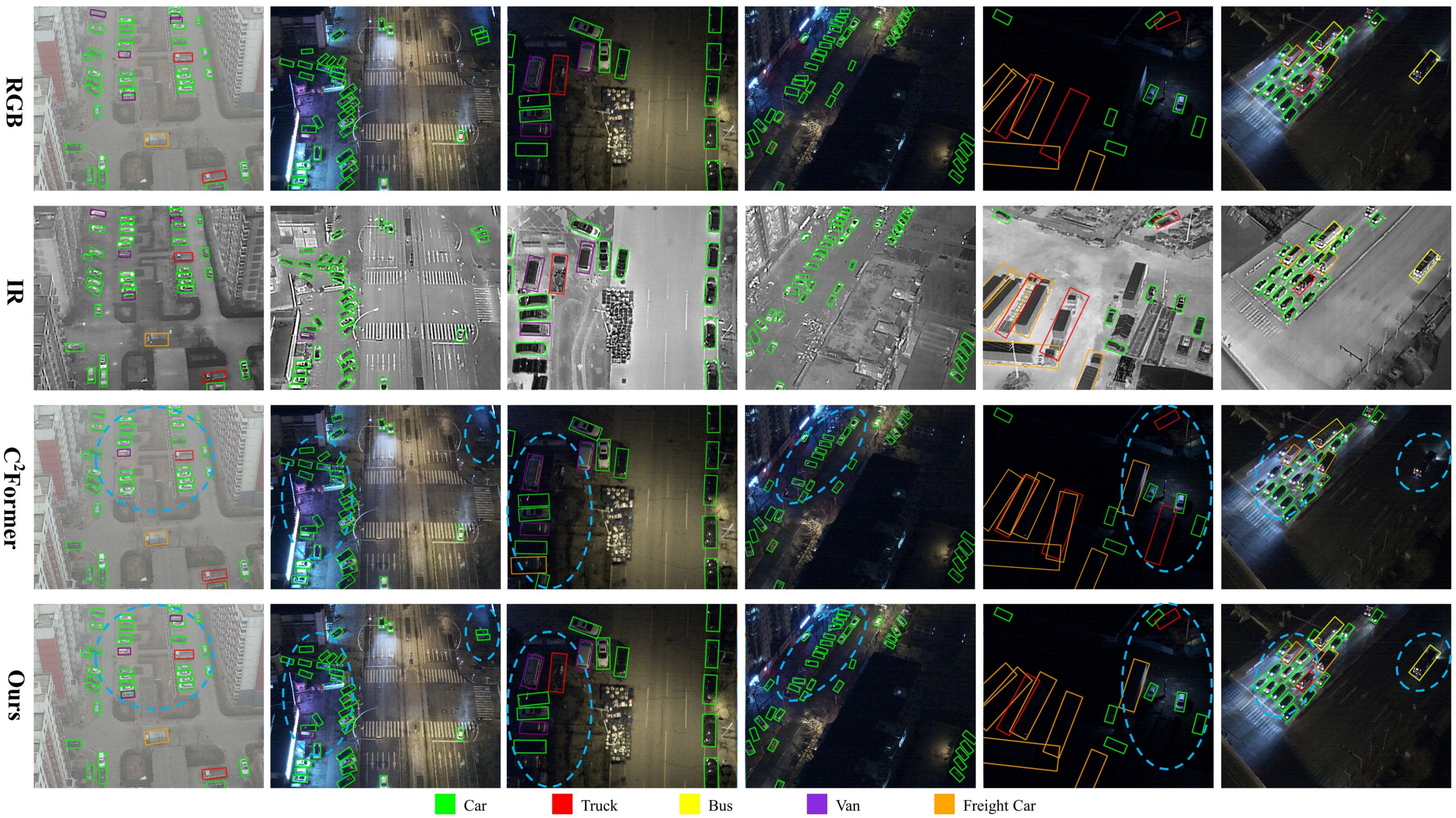}
    \vspace{-0.6cm}
    \caption{Visualization of prediction results on the DroneVehicle dataset, with a confidence threshold set to 0.5. The detection boxes in RGB and IR represent ground truths. The comparison of detection results within the blue dashed circles indicates that our method demonstrates superior visual performance for each category.}
    \label{fig:dv_vis}
    \vspace{-0.4cm}
\end{figure*}

\begin{table*}[!ht]
    \centering
    \setlength{\tabcolsep}{8pt}
    \caption{comprehensive comparative experiments on the droneVehicle dataset. we compared the dmm method with both single-modal and multi-spectral object detectors, all employing OBB detection heads. The best results are highlighted in bold, while the second-best results are underlined.}
    \label{tab:pare_dv}
\begin{tabular}{c|c|c|c|ccccc|c}
\toprule
\textbf{Method}                                 & \textbf{Basic Detector}    & \textbf{Type}  & \textbf{Modality}       & \textbf{Car}     & \textbf{Truck}   & \textbf{Freight Car} & \textbf{Bus}     & \textbf{Van}     & \textbf{mAP@0.5} \\ \midrule
RetinaNet~\cite{Lin_2017_ICCV}                  & -                          & One Stage      & \multirow{6}{*}{RGB}    & 78.5             & 34.4             & 24.1                 & 69.8             & 28.8             & 47.1             \\
R\textsuperscript{3}Det~\cite{yang2021r3det}    & -                          & One Stage      &                         & 80.3             & 56.1             & 42.7                 & 80.2             & 44.4             & 60.8             \\
S\textsuperscript{2}ANet~\cite{han2021align}    & -                          & One Stage      &                         & 80.0             & 54.2             & 42.2                 & 84.9             & 43.8             & 61.0             \\
Faster R-CNN~\cite{NIPS2015_14bfa6bb}           & -                          & Two Stage      &                         & 79.0             & 49.0             & 37.2                 & 77.0             & 37.0             & 55.9             \\
RoITransformer~\cite{ding2019learning}          & -                          & Two Stage      &                         & 61.6             & 55.1             & 42.3                 & 85.5             & 44.8             & 61.6             \\
Oriented R-CNN~\cite{Xie_2021_ICCV}             & -                          & Two Stage      &                         & 80.1             & 53.8             & 41.6                 & 85.4             & 43.3             & 60.8             \\ \midrule
RetinaNet~\cite{Lin_2017_ICCV}                  & -                          & One Stage      & \multirow{6}{*}{IR}     & 88.8             & 35.4             & 39.5                 & 76.5             & 32.1             & 54.5             \\
R\textsuperscript{3}Det~\cite{yang2021r3det}    & -                          & One Stage      &                         & 89.5             & 48.3             & 16.6                 & 87.1             & 39.9             & 62.3             \\
S\textsuperscript{2}ANet~\cite{han2021align}    & -                          & One Stage      &                         & 89.9             & 54.5             & 55.8                 & 88.9             & 48.4             & 67.5             \\
Faster R-CNN~\cite{NIPS2015_14bfa6bb}           & -                          & Two Stage      &                         & 89.4             & 53.5             & 48.3                 & 87.0             & 42.6             & 64.2             \\
RoITransformer~\cite{ding2019learning}          & -                          & Two Stage      &                         & 89.6             & 51.0             & 53.4                 & 88.9             & 44.5             & 65.5             \\
Oriented R-CNN~\cite{Xie_2021_ICCV}             & -                          & Two Stage      &                         & 89.8             & 57.4             & 53.1                 & 89.3             & 45.4             & 67.0             \\ \midrule
UA-CMDet~\cite{sun2022drone}                    & RoITransformer             & Two Stage      & \multirow{8}{*}{RGB+IR} & 87.5             & 60.7             & 46.8                 & 87.1             & 38.0             & 64.0             \\
Halfway Fusion~\cite{liu2016multispectral}      & Faster R-CNN               & Two Stage      &                         & 90.1             & 62.3             & 58.5                 & 89.1             & 49.8             & 70.0             \\
CIAN~\cite{zhang2019cross}                      & -                          & One Stage      &                         & 90.1             & 63.8             & 60.7                 & 89.1             & 50.3             & 70.8             \\
AR-CNN~\cite{zhang2021weakly}                   & Faster R-CNN               & Two Stage      &                         & 90.1             & 64.8             & 62.1                 & 89.4             & 51.5             & 71.6             \\
MBNet~\cite{zhou2020improving}                  & -                          & One Stage      &                         & 90.1             & 64.4             & 62.4                 & 88.8             & 53.6             & 71.9             \\
TSFADet~\cite{yuan2022translation}              & Oriented R-CNN             & Two Stage      &                         & 89.9             & 67.9             & 63.7                 & \underline{89.8} & 54.0             & 73.1             \\
TSFADet~\cite{yuan2022translation}              & Cacade R-CNN               & Two Stage      &                         & 90.0             & 69.2             & \underline{65.5}     & 89.7             & 55.2             & 73.9             \\
C\textsuperscript{2}Former~\cite{yuan2024c}     & S\textsuperscript{2}ANet   & One Stage      &                         & \underline{90.2} & 68.3             & 64.4                 & \underline{89.8} & 58.5             & 74.2             \\ \midrule
DMM(Ours)                                       & Faster R-CNN               & Two Stage      & \multirow{2}{*}{RGB+IR} & \textbf{90.4}    & \underline{77.8} & 63.0                 & 88.7             & \underline{66.0} & \underline{77.2} \\  
\textbf{DMM(Ours)}                              & S\textsuperscript{2}ANet   & One Stage      &                         & \textbf{90.4}    & \textbf{79.8}    & \textbf{68.2}        & \textbf{89.9}    & \textbf{68.6}    & \textbf{79.4}    \\ \bottomrule
\end{tabular}
\vspace{-0.4cm}
\end{table*}

{\bf Experiment Settings:} Our experiments were conducted on an NVIDIA 4090 GPU with 24GB of memory. The model's code environment is based on CUDA 11.6 and PyTorch 1.13.1, built upon the MMdetection and MMrotate frameworks. We employed pretrained VMamba as our backbone network, with input image dimensions set to 512x640 and a batch size of 2. The optimizer used is AdamW, with an initial learning rate of 0.0001 and a weight decay of 0.05. Data augmentation is limited to random flipping with a probability of 0.5. To enhance model stability, we removed all zero-area bounding boxes from the ground truths during training step. The model was trained for 12 epochs to ensure a fair comparison with previous methods. For the DroneVehicle dataset, we followed the settings from previous studies~\cite{sun2022drone,yuan2022translation,yuan2024c}, using the IR modality labels as ground truth. For the VEDAI dataset, we trained the model on all 9 categories, including those with fewer than 50 instances.

{\bf Metrics:} Mean Average Precision (mAP) is a widely used evaluation metric for object detection tasks. It assesses the precision of the detected bounding boxes by comparing them with the ground truth boxes based on the Intersection over Union (IoU) metric. mAP is the mean of the Average Precision (AP) values for all classes in the dataset, providing an overall performance measure for the object detection model across different object categories. In this work, we follow the common practice of using an IoU threshold of 0.5 to calculate the mAP metric.
\begin{figure*}[!htbp]
    \centering
    \includegraphics[width=1\linewidth]{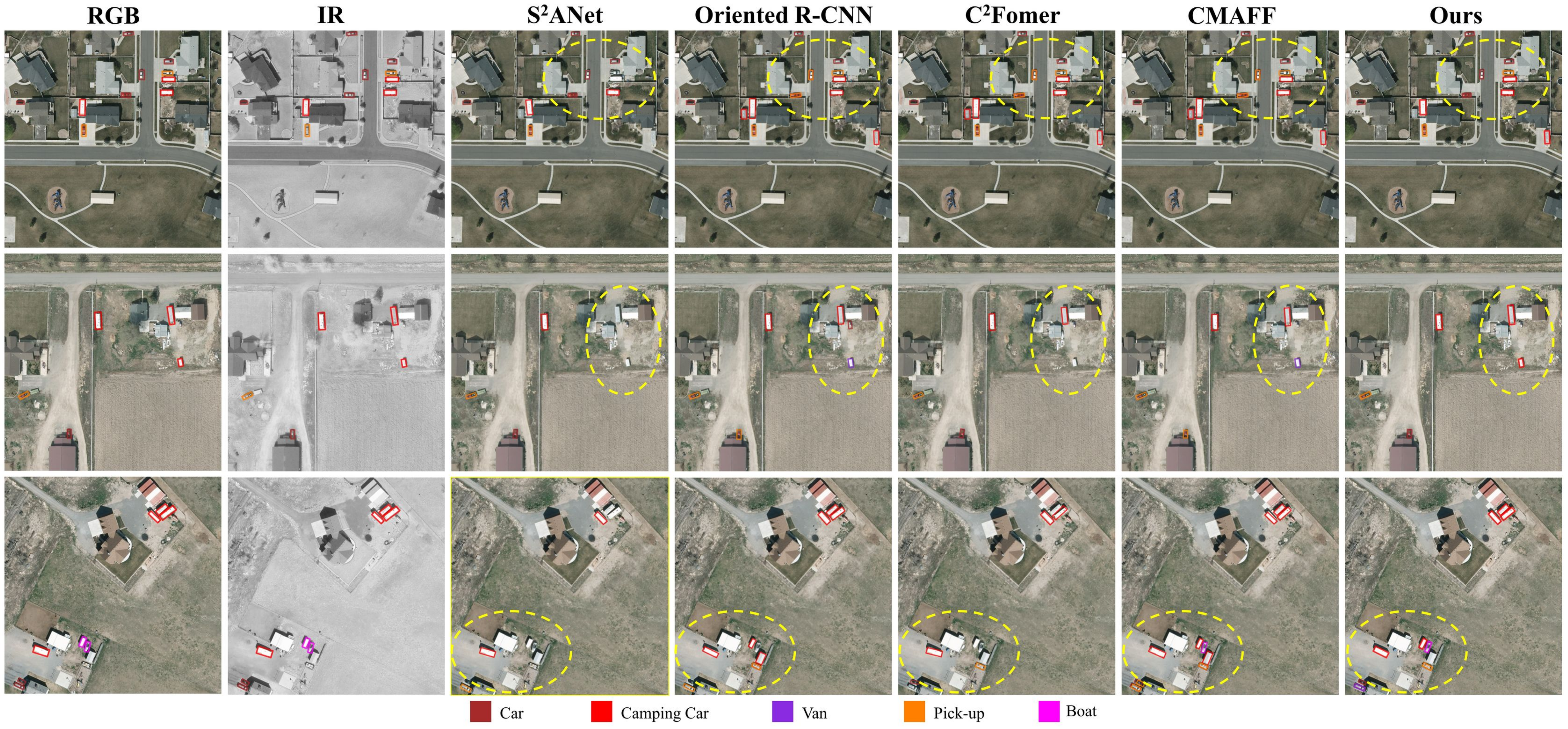}
    \vspace{-0.6cm}
    \caption{Visualization of prediction results on the VEDAI dataset, with a confidence threshold set to 0.5. The detection boxes in RGB and IR represent ground truths. The comparison of detection results within the yellow dashed circles demonstrates that our method has a distinct advantage in identifying small aerial targets.}
    \label{fig:vedai_vis}
    \vspace{-0.4cm}
\end{figure*}
\begin{table*}[htbp]
  \centering
  \setlength{\tabcolsep}{3pt}
  \caption{comprehensive comparative experiments on the VEDAI dataset. All the methods utilize OBB detection heads and the metrics are the averages from ten-fold cross-validation. The best results are highlighted in bold, while the second-best results are underlined.}
  \label{tab:pare_vedai}
    \begin{tabular}{c|c|c|ccccccccc|c}
    \toprule
    \textbf{Method}                                    & \textbf{Type} & \textbf{Modality}        & \textbf{Car}     & \textbf{Truck }  & \textbf{Tractor} & \textbf{Camping Car} & \textbf{Van} & \textbf{Pick-up} & \textbf{Boat} & \textbf{Plane} & \textbf{Others} & \textbf{mAP@0.5}       \\
    \midrule
    RetinaNet~\cite{Lin_2017_ICCV}                     & One Stage &\multirow{5}[2]{*}{RGB}    & 48.9             & 16.8             & 15.9             & 21.4             & 5.9              & 37.5             & 4.4              & 21.2             & 14.1             & 20.7             \\
    S\textsuperscript{2}ANet~\cite{han2021align}       & One Stage &                           & 74.5             & 47.3             & 55.6             & 61.7             & 32.5             & 65.1             & 16.7             & 7.1              & 39.8             & 44.5             \\
    Faster R-CNN~\cite{NIPS2015_14bfa6bb}              & Two Stage &                           & 71.4             & 54.2             & 61.0             & 70.5             & 59.5             & 67.6             & 52.3             & 77.1             & 40.1             & 61.5             \\
    RoITransformer~\cite{ding2019learning}             & Two Stage &                           & 77.3             & 56.1             & 64.7             & 73.6             & 60.2             & 71.5             & 56.7             & \underline{85.7} & 42.8             & 65.4             \\
    Oriented R-CNN~\cite{Xie_2021_ICCV}                & Two Stage &                           & 77.6             & \underline{59.7} & 62.8             & 76.7             & 60.9             & 72.3             & 60.1             & 84.0             & 43.6             & 66.4             \\
    \midrule
    RetinaNet~\cite{Lin_2017_ICCV}                     & One Stage &\multirow{5}[2]{*}{IR}     & 44.2             & 15.3             & 9.4              & 17.1             & 7.2              & 32.1             & 4.0              & 33.4             & 5.7              & 18.7             \\
    S\textsuperscript{2}ANet~\cite{han2021align}       & One Stage &                           & 73.0             & 39.2             & 41.9             & 59.2             & 32.3             & 65.6             & 13.9             & 12.0             & 23.1             & 40.0             \\
    Faster R-CNN~\cite{NIPS2015_14bfa6bb}              & Two Stage &                           & 71.6             & 49.1             & 49.2             & 68.1             & 57.0             & 66.5             & 35.6             & 71.6             & 29.5             & 55.4             \\
    RoITransformer~\cite{ding2019learning}             & Two Stage &                           & 76.1             & 51.7             & 51.9             & 71.2             & 64.3             & 70.7             & 46.9             & 83.3             & 28.3             & 60.5             \\
    Oriented R-CNN~\cite{Xie_2021_ICCV}                & Two Stage &                           & 77.0             & 55.0             & 47.5             & 73.6             & 63.2             & 72.2             & 49.4             & 79.6             & 30.5             & 60.9             \\
    \midrule
    C\textsuperscript{2}Former+S\textsuperscript{2}ANet~\cite{yuan2024c} & One Stage &\multirow{4}[2]{*}{RGB+IR} & 76.7             & 52.0             & 59.8             & 63.2             & 48.0             & 68.7             & 43.3             & 47.0             & 41.9             & 55.6             \\
    DMM+S\textsuperscript{2}ANet(Ours)                 & One Stage &                           & 77.9             & 59.3             & 68.1             & 70.8             & 57.4             & 75.8             & 61.2             & 77.5             & 43.5             & 65.7             \\
    CMAFF+Oriented R-CNN~\cite{qingyun2022cross}        & Two Stage &                           & \underline{81.7} & 58.8             & \underline{68.7} & \underline{78.4} & \underline{68.5} & \underline{76.3} & \underline{66.0} & 72.7             & \underline{51.5} & \underline{69.2} \\
    \textbf{DMM+Oriented R-CNN(Ours)}        & Two Stage &                           & \textbf{84.2}    & \textbf{65.7}    & \textbf{72.3}    & \textbf{79.0}    & \textbf{72.5}    & \textbf{78.8}    & \textbf{72.3}    & \textbf{93.6}    & \textbf{56.2}    & \textbf{75.0}    \\
    \bottomrule
    \end{tabular}%
    \vspace{-0.4cm}
\end{table*}%

\subsection{Comparisons with State-of-the-Art Methods}
\textbf{Comparisons on DroneVehicle Dataset. } Table ~\ref{tab:pare_dv} presents the comparative results of our method against competing approaches. The compared methods include classic object detection approaches based on single-modal RGB and IR data: RetinaNet, R3Net, and S\textsuperscript{2}ANet for one-stage methods; Faster R-CNN, RoITransformer, and Oriented R-CNN for two-stage methods. Additionally, it includes multi-modal fusion detection methods such as Halfway Fusion, AR-CNN, CIAN, MBNet, TSFADet, and C\textsuperscript{2}Former. Among all the three single-modal methods and seven multi-modal methods, most of multi-modal fusion detection methods significantly outperform single-modal detectors, demonstrating the substantial advantage of multi-modal data over single-modal data.

In the previous multi-modal detectors, C\textsuperscript{2}Former achieves the highest detection accuracy with 74.2\% mAP@0.5, followed by TSFADet with 73.9\% mAP@0.5. By contrast, our method achieves the highest mAP@0.5 of 79.4\%, surpassing the SOTA by 5.2 points. Furthermore, when integrating our proposed modules into other classic detector like Faster R-CNN, we also observed superior performance exceeding the current SOTA. Additionally, the table reveals that the performance improvement of our method is primarily due to enhanced detection capabilities for categories with fewer instances, such as Truck (8657), Freight-Car (5064), and Van (4282). We provide visual comparisons of detection results on DroneVehicle dataset in Fig.~\ref{fig:dv_vis}. As illustrated by the blue dashed circles in the figure, our method identifies more instances with greater accuracy in challenging scenarios such as foggy weather and nighttime compared to SOTA methods, which aligns with our experimental observations. These comprehensive comparative experiments demonstrate the effectiveness and generalization capability of our method.

\begin{table}[!htbp]
  \centering
  \caption{Ablation experiments on the dronevehicle dataset. F represents Faster R-CNN and S represents S\textsuperscript{2}ANet.}
  \label{tab:ablation}
  \setlength{\tabcolsep}{1pt}
  \renewcommand\arraystretch{1.0}
    \begin{tabular}{c|cc|cccc|c}
    \toprule
    \multirow{2}[4]{*}{Method} & \multicolumn{2}{c|}{Detector} & \multicolumn{4}{c|}{Module}   & \multirow{2}[4]{*}{mAP@0.5} \\
\cmidrule{2-7}          & F & S & Mamba & DCFM  & MTA   & TPA   &  \\
    \midrule
    Baseline(B)                & \checkmark     & -              & -              & -              & -              & -              & 70.5 \\
    B+Mamba                    & \checkmark     & -              & \checkmark     & -              & -              & -              & 75.8 \\
    B+Mamba+DCFM               & \checkmark     & -              & \checkmark     & \checkmark     & -              & -              & 76.3 \\
    B+Mamba+MTA                & \checkmark     & -              & \checkmark     & -              & \checkmark     & -              & 75.6 \\
    B+Mamba+MTA+TPA            & \checkmark     & -              & \checkmark     & -              & \checkmark     & \checkmark     & 76.1 \\  \midrule
    \textbf{DMM+Faster R-CNN}  & \checkmark     & -              & \checkmark     & \checkmark     & \checkmark     & \checkmark     & 77.2 \\
    \textbf{DMM+S\textsuperscript{2}ANet}        & -              & \checkmark     & \checkmark     & \checkmark     & \checkmark     & \checkmark     & \textbf{79.4} \\
    \bottomrule
    \end{tabular}%
\end{table}%
\begin{table}[!htbp]
  \centering
  \setlength{\tabcolsep}{1pt}
  \caption{Comparison of our DMM method with the latest SOTA method C\textsuperscript{2}Former in terms of the number of model parameters, the maximum image size that can be processed during inference on a 24GB NVIDIA 4090 GPU, and detection performance.}
  \label{tab:cost}
    \begin{tabular}{c|c|c|c}
    \toprule
          & PARAMS & MAXIMUM IUPUT SIZE& mAP \\
    \midrule
    C\textsuperscript{2}Former & 118.47M & 1092x1092 & 74.2 \\
    DMM   & 87.97M(30.5M↓) & 3016x3016(762.8\%↑) & 79.4(5.2↑) \\
    \bottomrule
    \end{tabular}
\end{table}

\textbf{Comparisons on VEDAI Dataset. } We integrated our DMM into S\textsuperscript{2}A-Net and Oriented R-CNN, Table ~\ref{tab:pare_vedai} presents our experimental results on the VEDAI dataset and Fig.~\ref{fig:vedai_vis} provide visual comparisons of detection results. Currently, mainstream multimodal fusion detection research on the VEDAI dataset is primarily based on Horizontal Bounding Box (HBB), with limited exploration of OBB. Therefore, we mainly compare our approach with single-modal detectors on the VEDAI dataset, such as the one-stage methods RetinaNet and S\textsuperscript{2}A-Net, and the two-stage methods Faster R-CNN, RoITransfomer, and Oriented R-CNN. For multimodal methods, we compared with the convolutional attention-based method CMAFF and the SOTA method C\textsuperscript{2}Former on the DroneVehicle dataset. 

From the experimental results, it can be observed that two-stage methods significantly outperform one-stage methods. This is primarily because the instance sizes in the VEDAI dataset are very small, and except for several vehicle categories, other categories are inconsistently distributed across different fold splits, making it challenging for one-stage methods to directly regress to the precise location of the targets. Consequently, C\textsuperscript{2}Former achieved only 55.6\% mAP@0.5 on the VEDAI dataset, whereas our DMM+S\textsuperscript{2}A-Net achieved 65.7\%, an improvement of 10.1 percentage points, even surpassing most two-stage single-modal methods. Additionally, the CMAFF method based on Oriented R-CNN achieved 69.5\% mAP@0.5, higher than all single-modal methods, while our DMM+Oriented R-CNN exceeded it by 5.8 percentage points, reaching the optimal detection performance. The experimental results fully demonstrate the generality of our method for oriented object detection tasks in remote sensing.

\subsection{Ablation Study} 
We conducted our ablation experiments on the DroneVehicle dataset. To eliminate the influence of different detectors, we adopted the classic Faster R-CNN as the baseline and 
\begin{figure}[!htbp]
    \centering
    \includegraphics[width=1\linewidth]{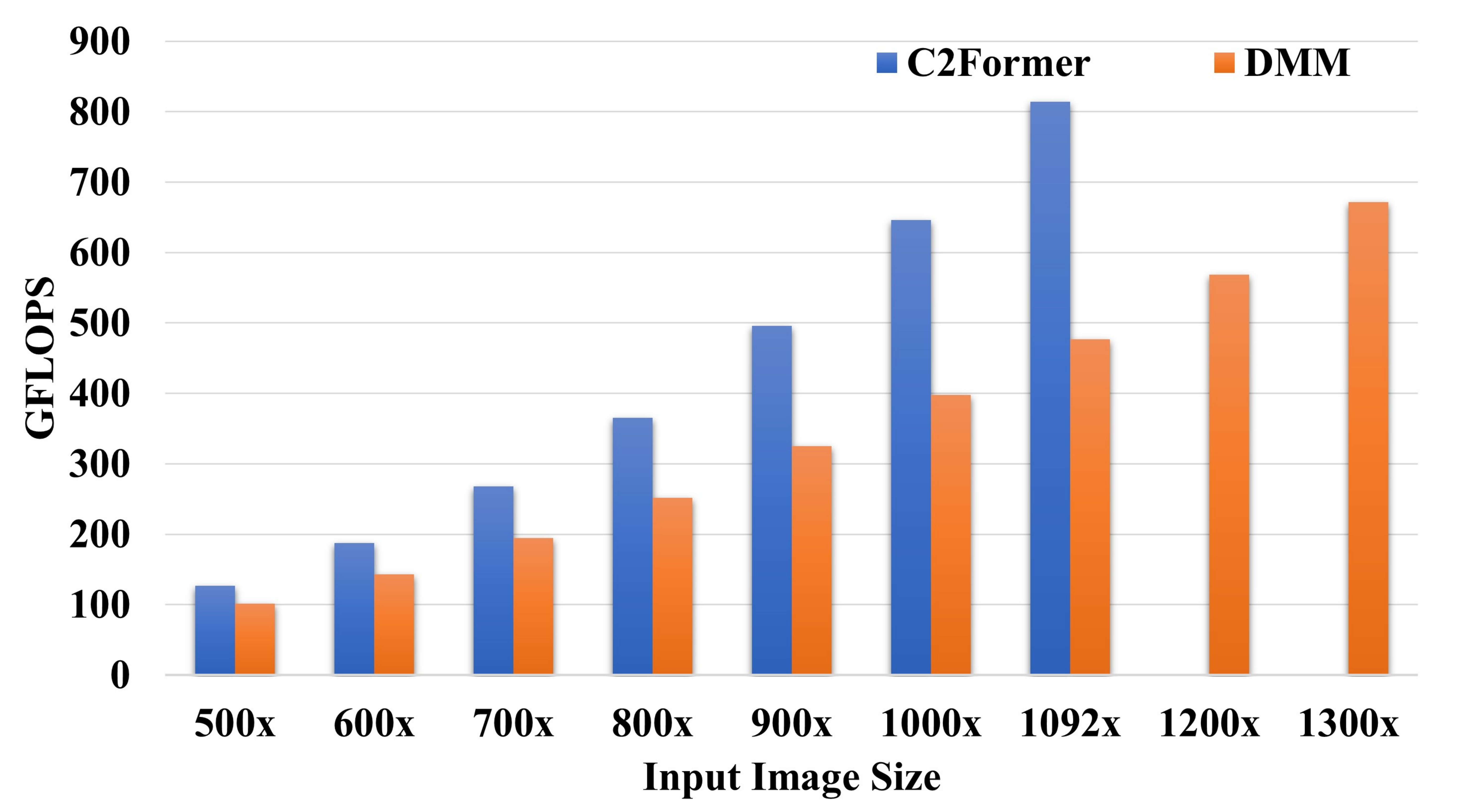}
    \caption{Comparison of GFLOPS for the model with different input image sizes. The input images have the same width and height, so only one dimension is marked on the horizontal axis. Due to the quadratic complexity of C\textsuperscript{2}Former, its memory usage exceeds physical limits for input sizes larger than 1100x1100, hence they are not shown.}
    \label{fig:flops}
\end{figure}
modified it into a dual-stream oriented detector. As shown in Table \ref{tab:ablation}, incorporating the Mamba model resulted in a 5.3\% increase in mAP. Adding the DCFM module further increased the mAP by 0.5\%, demonstrating the effectiveness of the Mamba model in processing remote sensing images. However, directly using the MTA module led to a 0.2\% decrease in mAP. In contrast, introducing the TPA auxiliary task raised the mAP to 76.1\%. Integrating all the modules, the model achieved its best performance with an mAP of 77.2\%. Additionally, the design of the detection head also influences model performance. Combining DMM with the high-performance S\textsuperscript{2}A-Net further improved the mAP to 79.4\%. These results indicate that each component contributes to the overall improvement in mAP.

\subsection{Comparisons of Computational Costs}
Table ~\ref{tab:cost} presents a comparison between our method and the previous SOTA method C\textsuperscript{2}Former. Both approaches use S\textsuperscript{2}ANet as the base detector and adhere to the same configuration. However, our model not only maintains a lower parameter count, with just 87.97M parameters, reducing approximately 25\%, but also achieves the highest detection performance. Moreover, as shown in Fig.~\ref{fig:flops}, with increasing input image sizes, the GFLOPS of our model increases almost linearly. In contrast, C\textsuperscript{2}Former's GFLOPS, though initially close to our model for smaller image sizes, significantly surpasses ours for larger images due to the quadratic complexity of the Transformer. Additionally, with larger input image sizes, the memory usage of C\textsuperscript{2}Former increases significantly. On a 24GB NVIDIA 4090 GPU, C\textsuperscript{2}Former supports a maximum inference size of only 1092x1092, whereas our model can handle inputs up to 3016x3016, which is approximately 7.6 times larger. These experimental results demonstrate the superior capability 
\begin{figure}[!htp]
    \centering
    \includegraphics[width=1\linewidth]{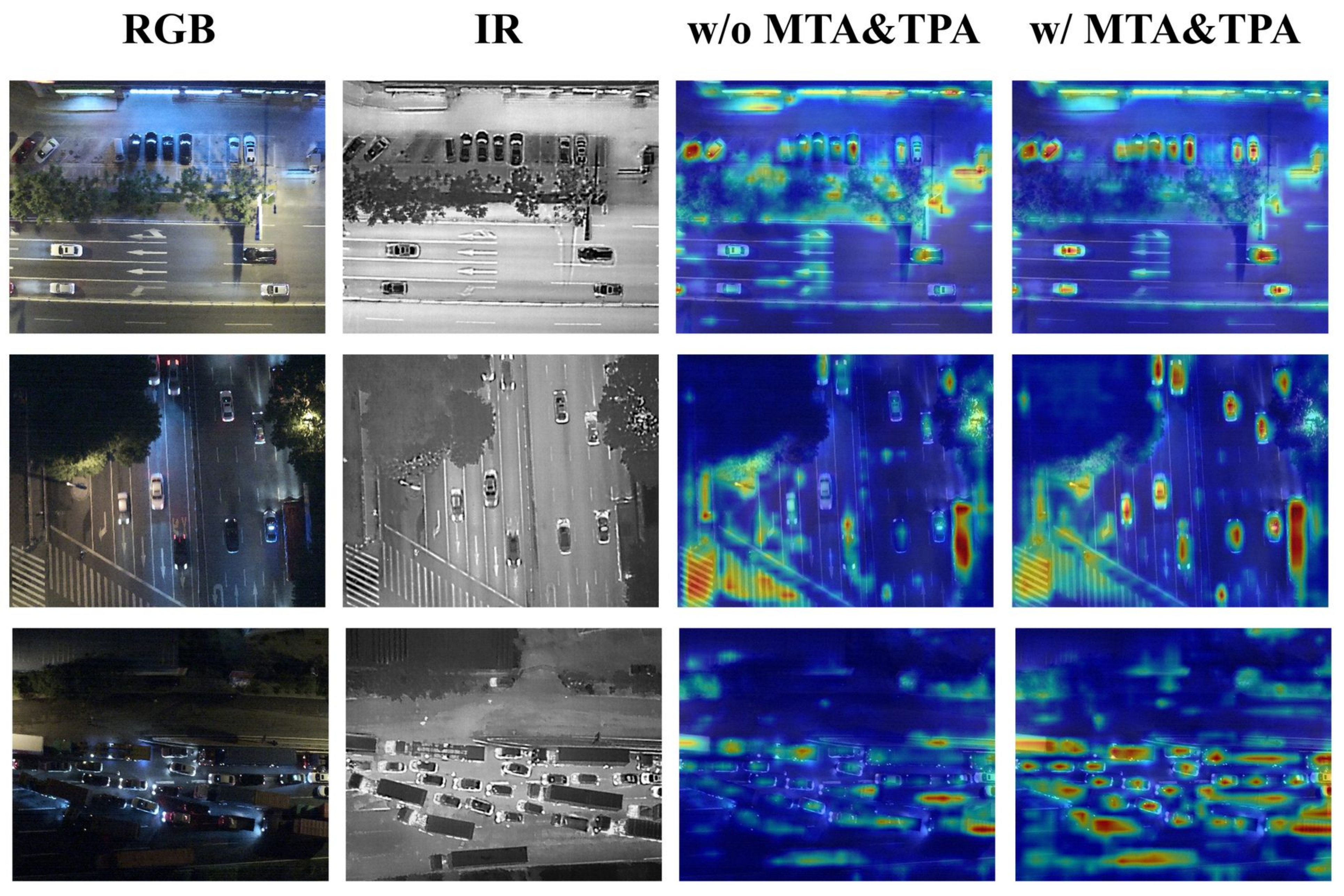}
    \vspace{-0.8cm}
    \caption{Visualization of intermediate feature maps. Comparison of the fused features with and without the use of the MTA and TPA modules.}
    \label{fig:feat_vis}
\end{figure}
of our model in handling high-resolution images in the remote sensing domain.

\subsection{Feature Analysis}
To validate the effectiveness of our mothod in enhancing detection performance, we present a comparative visualization of intermediate feature maps in Fig.~\ref{fig:feat_vis}. The comparisons show feature maps with and without these modules. It is evident that with the incorporation of the MTA and TPA modules, the model exhibits a higher degree of focus on the target objects while effectively suppressing background noise. 

To quantitatively demonstrate the effectiveness, we propose a metric called Spatial Feature Attention Contrast (\(SFAC\)). SFAC measures the model's focus on target features by calculating the ratio of the mean attention values within the target bounding boxes to the mean attention values of the background. Specifically, for an attention map, we first upsample it to match the original image size and then linearly scale the feature values to the range of 0 to 255. Finally, we compute \(SFAC\) using the following formula:
\begin{equation}
    \begin{split}
        \begin{aligned}
            SFAC &= \frac{1}{n}\sum_{k=1}^n{\frac{ \sum_{(i,j) \in \textbf{P}_{\text{bbox}}^k} \textbf{I}(i,j) }{\sum_{(i,j)  \notin \textbf{P}_{\text{bbox}}^k} \textbf{I}(i,j)}}
        \end{aligned}
    \end{split}
\end{equation}
where \( n \) is the number of images in the test set, \( \textbf{I}(i,j) \) denotes the pixel value of the pixel \( (i, j)\) in the \( k \)-th image, \( \textbf{P}_{bbox}^k \) represents the set of pixels within the ground truth bounding boxes in the \( i \)-th image. The higher the SFAC value, the greater the model's focus on the target.

\begin{figure}[!htbp]
    \centering
    \includegraphics[width=1\linewidth]{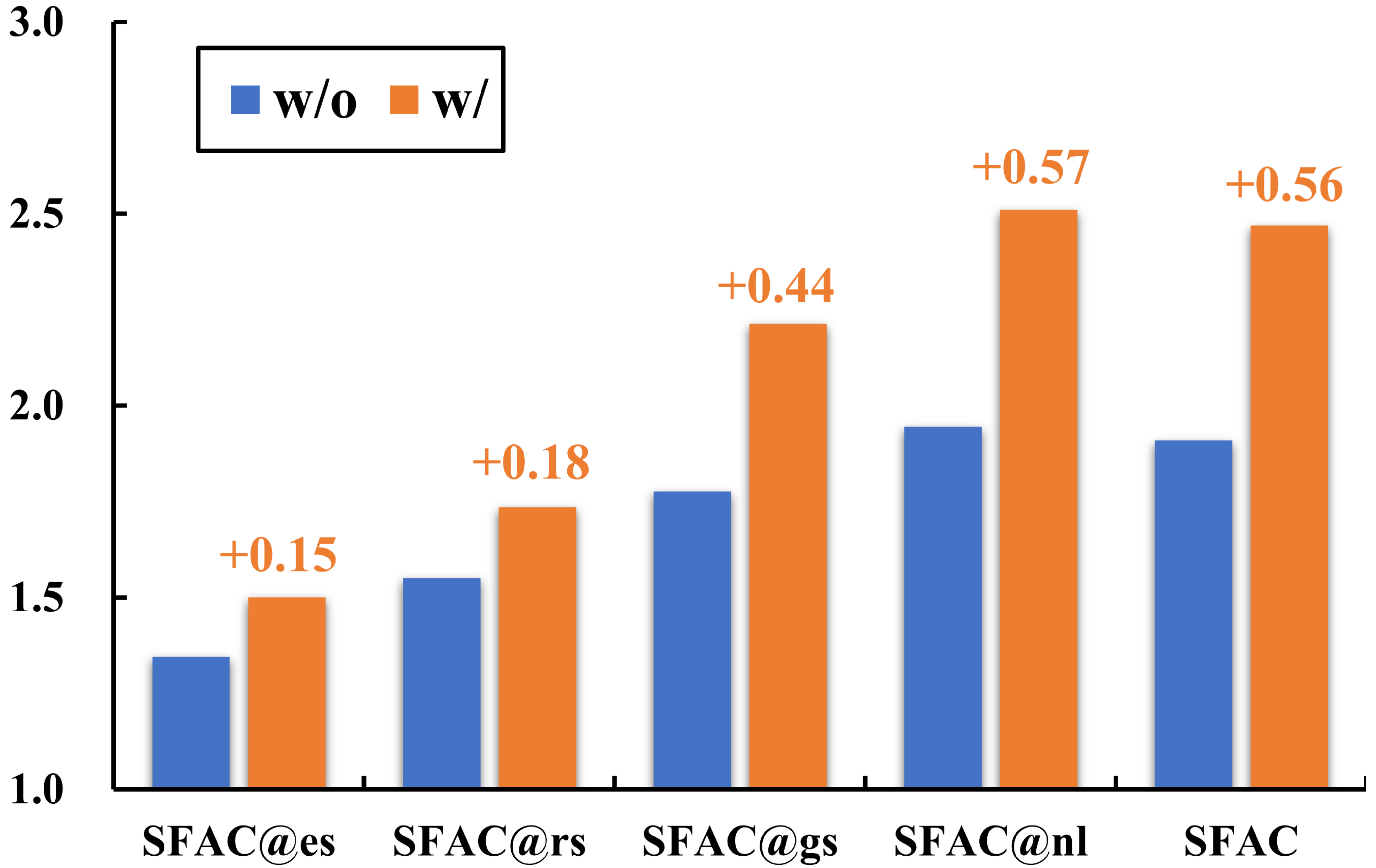}
    \caption{Comparison of \(SFAC\) metrics between the baseline and our method for targets of different sizes. \(es\), \(rs\), \(gs\), and \(nl\) represent extremely small, relatively small, generally small, and normal size, respectively. \(SFAC@es\) denotes the Spatial Feature Attention Contrast for all extremely small targets, with similar definitions for the other metrics.}
    \label{fig:quant}
\end{figure}

Based on the definitions in ~\cite{10168277}, we calculate the \(SFAC\) for different target sizes: extremely small targets with an area less than or equal to 144 pixels, relatively small targets with an area greater than 144 pixels but less than or equal to 400 pixels, generally small targets with an area greater than 400 pixels but less than or equal to 1024 pixels, normal targets with an area greater than 1024 pixels, and all targets combined. The results are shown in Fig.~\ref{fig:quant}. Quantitative results indicate that our method significantly enhances the model's focus on targets compared to background noise. Specifically, there are notable improvements for generally small targets and normal targets, with increases of 0.44 and 0.57, respectively. The attention toextremely small and relatively small targets also shows improvement. Overall, our method enhances the model's target perception capability by approximately 30\%. These quantitative results strongly demonstrate the effectiveness of our method.

\section{Conclusions} \label{sec:conclusions}
In this work, we proposed Disparity-guided Multispectral Mamba (DMM), a framework for multispectral oriented object detection in remote sensing. Our DMM includes a DCFM module, which integrates global context information by leveraging modality disparity information as guidance to adaptively resolve inter-modal conflicts, thereby achieving efficient cross-modal feature fusion. Additionally, we designed an MTA module to suppress noise and focus on target regions, minimizing the impact of intra-modal variations on subsequent fusion. To ensure the effectiveness of the MTA module, we applied a TPA auxiliary task, using single-modal losses as penalty terms to constrain the optimization process of the model. Extensive experiments on two challenging datasets demonstrate the generalization capability of DMM, achieving SOTA results. In the future, we will explore the application of our method in a wider range of scenarios.

\bibliographystyle{IEEEtran}

\bibliography{IEEEabrv, ref.bib}
\begin{IEEEbiographynophoto}
{Minghang Zhou}
is currently pursuing the M.S. degree with the School of Computer Science and Engineering, University of Electronic Science and Technology of China, Chengdu, China. His research interests include computer vision and remote sensing image processing.
\end{IEEEbiographynophoto}
\begin{IEEEbiographynophoto}
{Tianyu Li}
received his Ph.D. degree from the School of Electrical and Computer Engineering at Purdue University, USA, in 2023. He is currently a postdoctoral researcher at the Center for Future Media, University of Electronic Science and Technology of China, located in Chengdu, China. His research focuses on image processing using statistical models and deep learning models, as well as time series forecasting and anomaly detection in time series.
\end{IEEEbiographynophoto}
\begin{IEEEbiographynophoto}
{Chaofan Qiao}
is currently pursuing the Ph.D. degree with the School of Computer Science and Engineering, University of Electronic Science and Technology of China, Chengdu, China. His major research interests include multimedia, computer vision, and machine learning.
\end{IEEEbiographynophoto}
\begin{IEEEbiographynophoto}
{Dongyu Xie}
is currently pursuing the M.S. degree with the School of Computer Science and Engineering, University of Electronic Science and Technology of China, Chengdu, China. His research interests include multimedia, computer vision, and machine learning.
\end{IEEEbiographynophoto}
\begin{IEEEbiographynophoto}{Guoqing Wang} \textbf{(Member, IEEE)} received the Ph.D. degree from The University of New South Wales, Australia, in 2021. He is currently with the School of Computer Science and Engineering, University of Electronic of Science and Technology of China. He has authored and co-authored more than 40 scientific articles at top venues, including IJCV, IEEE TIP, IEEE TIFS, ICCV, ACM MM, etc. His research work at UNSW has been recognized as the Australian Dean’s Award for Outstanding Ph.D. Theses. His research interests include machine learning and unmanned system, with special emphasis on cognition and embodied agents.
\end{IEEEbiographynophoto}
\begin{IEEEbiographynophoto}
{Ningjuan Ruan} received the Ph.D. degree in Electronic Science and Technology from Nanjing University in March 2019. She is currently the Director of the Advanced Optical Remote Sensing Beijing Key Laboratory. She is also a member of the Expert Committee of the Core Technology Integration Platform at Nanjing University, under the Ministry of Education, and a member of the Academic Committee of the Key Laboratory of Space Optoelectronic Detection and Perception, Ministry of Industry and Information Technology. She is affiliated with the Beijing Institute of Space Mechanics and Electricity, Beijing 100094, China.
\end{IEEEbiographynophoto}
\begin{IEEEbiographynophoto}
{Lin Mei} received the Ph.D. degree from Xi’an Jiaotong University, Xi’an, China, in 2000. He is a researcher at Donghai Laboratory. His research interests include computer vision, artificial intelligence, internet of things, big data, etc.
\end{IEEEbiographynophoto}
\begin{IEEEbiographynophoto}
{Yang Yang}\textbf{(Senior Member, IEEE)} received the bachelor’s degree from Jilin University, Changchun, China, in 2006, the master’s degree from Peking University, Beijing, China, in 2009, and the Ph.D. degree from The University of Queensland, Brisbane, QLD, Australia, in 2012, all in computer science. He is currently with the University of Electronic Science and Technology of China, Chengdu, China. His current research interests include multimedia content analysis, computer vision, and social media analytics.
\end{IEEEbiographynophoto}

\end{document}